\documentclass[11pt,a4paper]{article}
\usepackage[T1]{fontenc}
\usepackage[utf8]{inputenc}
\usepackage{authblk}
\usepackage[square,numbers]{natbib}
\usepackage{hyperref}

\usepackage{array,multirow,graphicx}
\usepackage{mathtools}

\usepackage{pdfpages}
\usepackage{algorithm}
\usepackage{algorithmic}
\usepackage{adjustbox}
\usepackage{multirow}
\usepackage{caption}
\usepackage[utf8]{inputenc}
\usepackage{booktabs}
\usepackage{balance}
\usepackage{pgfplots}
\usepackage{graphicx}
\usepackage{caption}
\usepackage[subrefformat=parens,labelformat=parens]{subcaption}
\usepackage{pgfplots, pgfplotstable}
\pgfplotsset{compat=1.14}
\usetikzlibrary{pgfplots.groupplots,shapes,decorations}

\usepackage[margin=1.0in]{geometry}

\usepackage{fancyhdr}

\usepackage[square,numbers]{natbib}

\title{Efficient Malware Analysis Using Metric Embeddings}

\author[1]{Ethan M. Rudd\thanks{ethan.rudd@mandiant.com}}
\author[1]{David Krisiloff\thanks{david.krisiloff@mandiant.com}}
\author[1]{Scott Coull\thanks{scott.coull@mandiant.com}}
\author[2]{Daniel Olszewski\thanks{dolszewski@ufl.edu}}
\author[3,4]{Edward Raff\thanks{raff\_edward@bah.com}}
\author[4]{James Holt\thanks{holt@lps.umd.edu}}
\affil[1]{Mandiant Inc.}
\affil[2]{University of Florida}
\affil[3]{Booz Allen Hamilton}
\affil[4]{Laboratory for Physical Sciences, University of Maryland}
\date{} 

\begin{document}

\maketitle

\begin{abstract}

% new abstract <= 200 words 
In this paper, we explore the use of metric learning to embed Windows PE files in a low-dimensional vector space for downstream use in a variety of applications, including malware detection, family classification, and malware attribute tagging. 

Specifically, we enrich labeling on malicious and benign PE files using computationally expensive, disassembly-based malicious capabilities. Using these capabilities, we derive several different types of metric embeddings utilizing an embedding neural network trained via contrastive loss, Spearman rank correlation, and combinations thereof. 

We then examine performance on a variety of transfer tasks performed on the EMBER and SOREL datasets, demonstrating that for several tasks, low-dimensional, computationally efficient metric embeddings maintain performance with little decay, which offers the potential to quickly retrain for a variety of transfer tasks at significantly reduced storage overhead. We conclude with an examination of practical considerations for the use of our proposed embedding approach, such as robustness to adversarial evasion and introduction of task-specific auxiliary objectives to improve performance on mission critical tasks. 

\end{abstract}

\section{Introduction}

Malware analysis is a complex process involving highly skilled experts and many person-hours. Given the number of new files seen each day (more than 500,000 on VirusTotal alone \cite{vtstats}) automation of malware analysis is a necessity. Development of new analysis tools provides an avenue for more efficient malware analysis teams.

Fortunately, malware analysis tasks are often amenable to machine learning (ML) solutions. The tasks (e.g., malware detection or malware family classification) are complex enough that traditional rules-based approaches remain brittle and require frequent updating. At the same time, it is possible to acquire and label large data sets to train ML models using threat feeds and crowdsourcing services, like VirusTotal or Reversing Labs.

One notable downside, however, is that ML models come with significant technical debt: they need to be retrained as malware evolves and often interdependencies between various model components can be hard to understand or predict \cite{sculley2014machine}. Even with a consistent feature vector representation, when training on industry-scale datasets, feature stores may require tens of terabytes and model re-training can take multiple weeks.

This motivates the question: what if we can use ML to derive low-dimensional representations which capture semantic behavior of malware/goodware that can be used to speed up and reduce resource requirements for downstream tasks? This could significantly enhance capability for training classifiers for novel applications, performing rapid iteration/experimentation, and efficiently updating deployed models, all while reduced processing and storage requirements.

Since other applications of applied ML, including biometrics and information retrieval (IR) systems, have utilized metric learning to derive low-dimensional embeddings for similar downstream tasks, in this paper, we explore whether we can apply metric learning in a similar vein towards various malware analysis-oriented ML tasks. Using metric embeddings, we aim to simplify some of the engineering costs associated with running and maintaining a suite of different downstream ML tooling. 

Contrary to other applications of ML classification, where data can trivially be assigned labels corresponding to one or more classes/attributes, labeling for malware analysis tasks can be more difficult \cite{251586}. Moreover, data for malware analysis often includes telemetry and metadata beyond hard labels which could ideally be used to enrich our metric embeddings. In this paper, we explore techniques to enrich our embeddings with complex semantic information provided by computationally-expensive tools (e.g., disassembly). This allows us to explore whether it is possible to approximate more expensive analysis with lower-overhead static representations. 

When generating our embeddings, we utilize Mandiant's CAPA tool; an open source tool, which utilizes rules and heuristics in conjunction with disassembly to yield  capability labels (e.g., file read/write, registry key generation, process creation, data send/receive over networks, socket connection, base64/XOR encoding) associated with a given PE, ELF, and .NET file as well as shellcode snippets. We enrich samples from the EMBER dataset with these computationally-expensive capability labels, and using these labels generate different types of efficient embeddings, including a Siamese embedding, which utilizes a contrastive loss over clusters of CAPA attributes, as well as a novel ranking embedding, which uses the Spearman rank correlation coefficient as a loss and aims to embed ranked degree of similarity between different CAPA attribute clusters. We then perform comparisons of these different metric embedding loss functions across two different datasets: EMBER and SOREL-20M, on three different downstream transfer tasks: malware detection, malware family classification and malware attribute classification, making comparisons to original dataset benchmarks where applicable. Finally, we perform an analysis of practical considerations surrounding the use of our metric embeddings, including robustness to adversarial evasion, qualitative analysis of the underlying learned metric space, and the use of task-specific auxiliary loss functions to improve performance on critical tasks.

The contributions of our paper are summarized as follows:
\begin{enumerate}
    \item We are the first to propose a metric learning approach that incorporates semantically-rich info into an efficiently-computable metric space for malware analysis tasks.
    \item We show a number of novel training regimes for the model using a Siamese network, including contrastive, Spearman, and task-specific losses. In doing so, we are the first to combine contrastive and Spearman losses to provide both coarse and fine-grained similarity information during training.
    \item We demonstrate that generic metric embeddings can successfully tackle several important malware analysis tasks, including detection, family classification, and type prediction.
    \item We evaluate the adversarial robustness of metric embeddings in the malware analysis problem space, which has not been previously explored.
\end{enumerate}

\section{Background and Related Work}
\label{sec:background}

Metric learning is a machine learning task that focuses on learning distances (metrics/measures) between objects that captures some semantically meaningful notion of similarity.  These learned metric functions play an important role in fields including information retrieval, ranking, and recommendation systems \cite{kaya2019deep}. The key property of a similarity metric/measure is that it maps similar objects close together and dissimilar objects far apart within the learned metric space. In practice, the objects are represented by a set of features, the metric function is the transformation of the features into a common metric space, and the learning process finds a transformation such that the similarity/dissimilarity behavior is correct with respect to the labels provided during training. Various learning architectures have been proposed, including Siamese networks that learn from the distances between pairs of objects, and triplet networks that use three samples to capture both similarity and dissimilarity to an anchor \cite{koch2015siamese}, \cite{hoffer2015deep}. There are also different loss functions for each architecture along with other subtle modifications of the learning process that can be applied to improve results (e.g., specialized algorithms for stochastic gradient descent \cite{kaya2019deep}).

To perform metric learning, we require a dataset of objects along with their similarities. Since our objects are binary portable executables (PEs) we need to define a meaningful notion of similarity among binaries. Binary similarity gauges the likeness between two binary files and can be defined in several different ways. Perhaps the cleanest definition is that two binaries are similar if they were compiled from the same source code or contain a large fraction of the same source code \cite{shin2015recognizing}, \cite{xu2017neural}. This definition is particularly useful from a malware reverse engineering standpoint: knowing that a file contains source code from a known piece of malware can significantly speed up analysis. However, this definition introduces problems when labeling a dataset of binary files built from unknown source code, which is the case in practical malware analysis settings.

To the best of our knowledge, ours is the first work to pursue a discriminative-style contrastive loss as our method of learning a general purpose feature representation for multiple downstream malware tasks. Prior methods generally fall into the category of file hashes or disassembly-based methods to extract high-level code representations. In both cases, these methods are not amenable to being feature vectors that can adjust to population change over time (i.e., frequent retraining) in our deployment scenario (i.e., low overhead).  Compression-based hashing approaches \cite{raff_lzjd_digest,raff_shwel} allow for similarity search in a computationally efficient manner. Similarly, digital forensic hashes that produce a hash-code of fixed or variable length can be used to calculate similarities \cite{Winter2013,Breitinger:2013:MNA:2496032.2497148,Breitinger2014,DBLP:journals/jdfsl/BreitingerRB14,Breitinger2013a,Lillis2017,Winter2013,Oliver2013}. These approaches are often fast and low-overhead.  However, no learning step occurs for either of these types of approaches, which prevents the method from generalizing to changes in the population of malware or being used for downstream malware analysis tasks. 

The other primary approach is code similarity measures based on the disassembly \cite{Ding:2016:KMA:2939672.2939719,Ding2019,Massarelli2018} or call-graph \cite{Li,Yang2021,Chandramohan:2016:BCC:2950290.2950350}, and uses a neural network to train an auto-regressive model that can produce a fixed-length representation. This allows adapting the model over time, but the reliance on at least disassembling the given executable limits our ability to deploy such representations. Disassembly itself is computationally demanding, and often error-prone, as many times a file will not yield accurate disassembly without unpacking or de-obfuscation. These processes may need to be done manually. Combined this makes the approach undesirable for our goals. In fact, by contrast, one of the primary aims of our approach is to try to imbue this rich semantic knowledge directly into the learned metric space, while relying only on lightweight static features to compute that representation.

\section{Approach}
\label{sec:approach}

\subsection{Overview}

\begin{figure*}[!h]
    \centering
    \includegraphics{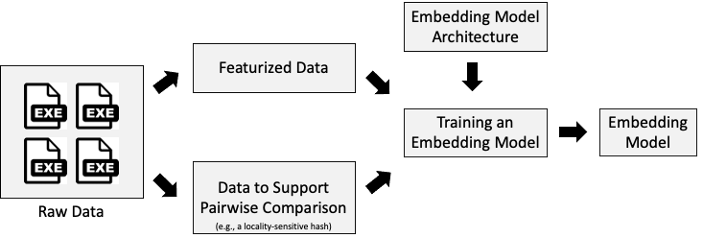}
    %\caption{System diagram representing the upstream training of an embedding model.}
    \caption{A system diagram depicting the upstream training of an embedding model. Training is conducted using a selected embedding model architecture, feature vectors from executable binaries, and additional data to support pairwise sample comparisons (e.g., CAPA labels).}
    \label{fig:upstream_training}
\end{figure*}

In this paper, we focus on building a model that produces embeddings of Windows PE files. The goal is to learn a representation that can be used for multiple downstream malware analysis tasks. The methodology can be separated into two phases.  Figure~\ref{fig:upstream_training} represents the first phase where an embedding model is trained.  To begin, we take the dataset of raw Windows PE binaries and apply two processes: (1) a featurization step, and (2) a step to compute file information that aids in determining the pairwise similarity between any two files.  For featurization, we focus on subject matter expert (SME)-derived, static features that are efficient to compute and which capture a broad range of malicious signals. These features include things such as the APIs imported, parsing errors, entropy, and byte distributions.

To determine pairwise similarity between two PE files, we use a tool like CAPA \cite{ballenthin2020capa} that detects capabilities of executable files (i.e., two files with overlapping capabilities are similar).  Notably, we hypothesize that using more complex similarity information than what is available naturally in the features (e.g., disassembly-based similarity vs. static features) will imbue the learned metric space with additional information without the added overhead of the more complex analysis techniques. Once the data and labels are defined, we specify an architecture for the embedding, which we instantiate with a multi-layered neural network.  An algorithm then takes those components as input and trains an embedding model.  The training algorithm can use a metric embedding network (e.g., a Siamese network) to learn the model parameters that effectively cause similar PE files to be near one another in the embedding space, and dissimilar files to be farther apart.

\begin{figure*}[!h]
    \centering
    \includegraphics{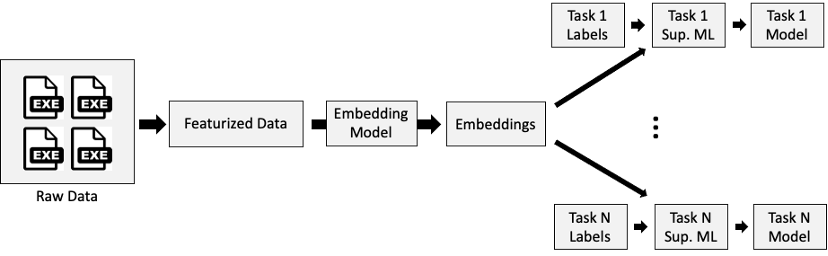}
    % \caption{System diagram representing the downstream use of the embedding model for multiple tasks.}
    \caption{A system diagram depicting the downstream use of a trained embedding model for multiple tasks. Features are extracted from executable binaries and fed as inputs to a trained embedding model, which generates a low-dimensional embedding representation of each of the input binaries. These embeddings along with auxiliary label information can be used for different downstream malware analysis tasks.}
    \label{fig:downstream_use}
\end{figure*}

During the second phase, as shown in Figure \ref{fig:downstream_use}, we measure the transferability of the embedding space to various malware analysis tasks. Concretely, we embed our training data and use that representation to train new models for each downstream task. By keeping the models used for the transfer process constant and only varying our embedding process we can make precise measurements of the utility of various similarity information, loss function, binary representation, or other parameters of our embedding network. The following sections detail our modeling setup.

\subsection{Network architecture}

% Each of the two binary representations (SME-derived features and raw bytes) has a unique network architecture.

\begin{figure}[!h]
    \centering
    \includegraphics[width=\linewidth]{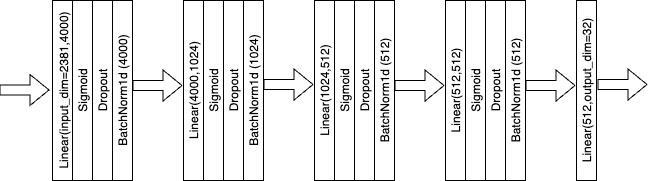}
    \caption{Architectural schematic of our embedding network.}
    \label{fig:net_architecture}
\end{figure}

Our embedding neural network architecture is shown in Figure \ref{fig:net_architecture}. The network takes as input 2381-dimensional static features defined by the EMBER 2018 dataset \cite{anderson2018ember}, though the approach is flexible to any static features used. The features are first normalized via standard scaling with respect to mean and variance on the EMBER 2018 training set, then fed to an embedding network, which is comprised of four dense layers with sigmoid activations of dimension 4,000, 1,024, 512, and 512. Between the layers we include both a BatchNorm and a Dropout layer with a dropout probability of 10\%. Following those layers is the final embedding layer using a linear activation with specified output dimension; in the majority of our experiments, we utilize an output dimension of 32 unless otherwise noted. During training the network outputs are then optionally normalized and losses, which compare CAPA attribute information, are evaluated and minimized via backpropagation.

% Optionally, e.g., if using using cosine distance, we could normalize over the final embedding layer.

%output dimension varies by experiment. 
%In some experiments we utilize cosine distance instead of Euclidean and for those cases we normalize the final embedding layer.  

% The network architecture for the raw binary file representation is based on the MalConv network. In our initial experiments we start with MalConv v1, but switch to MalConv v2 for its improved performance, including significant improvements in training time \cite{raff2018malware,raff2020classifying}. For the v1 architecture we add an additional dense layer with a linear activation for the final embedding. For the v2 architecture we explore using the penultimate layer, the layer before classification, as the embedding representation. A full exploration of the MalConv hyperparameter space and usage of additional dense layers is left for future work.

% As described below, we occasionally may include an additional term in the loss function to predict goodware or malware labels. When we do so, we add a single new prediction head with a single dense layer and sigmoid activation connected to the final embedding to make that prediction.

\subsection{Enriching Metric Embeddings with CAPA Labels}

The CAPA system detects various capabilities of a binary file using both the static analysis and disassembly and yields a set of capabilities for each file. These capabilities are categorical and non-mutually exclusive and are labeled with short text snippets (e.g., ``encode data using Base64"). We incorporate these generated sets of capabilities to enrich our embeddings via two different loss functions, which we apply both solo and in tandem (via summation) in our experiments.

\begin{figure}[!h]
\includegraphics[width=\linewidth]{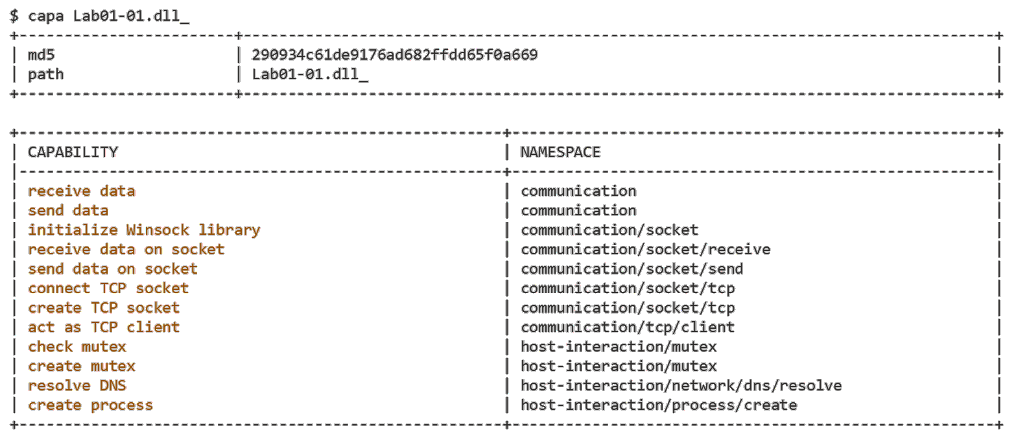}
\caption{A Sample CAPA report for a PE file \text{Lab01-01.dll\_} from \cite{ballenthin2020capa}.}
\end{figure}

\begin{figure}[!h]
    \centering
    \subfloat[Contrastive]{\label{fig:contrastiveCartoon}
            \includegraphics[width=0.33\linewidth]{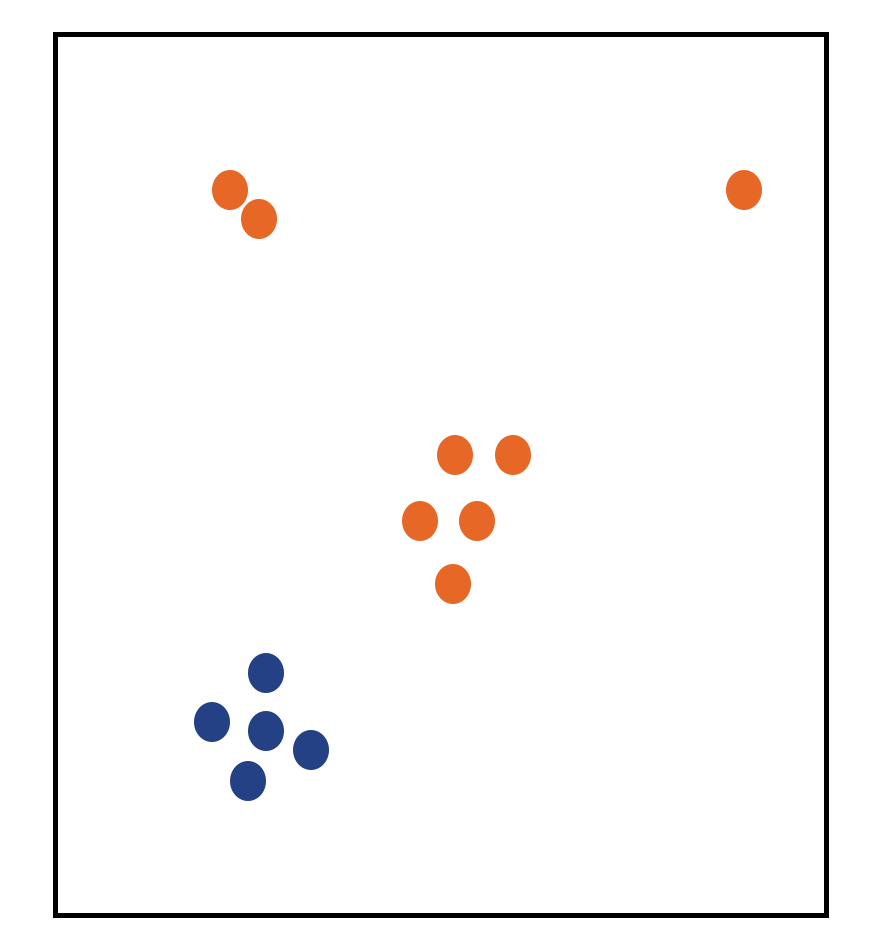}
    }
    \subfloat[Spearman]{\label{fig:SpearmanCartoon}
            \includegraphics[width=0.33\linewidth]{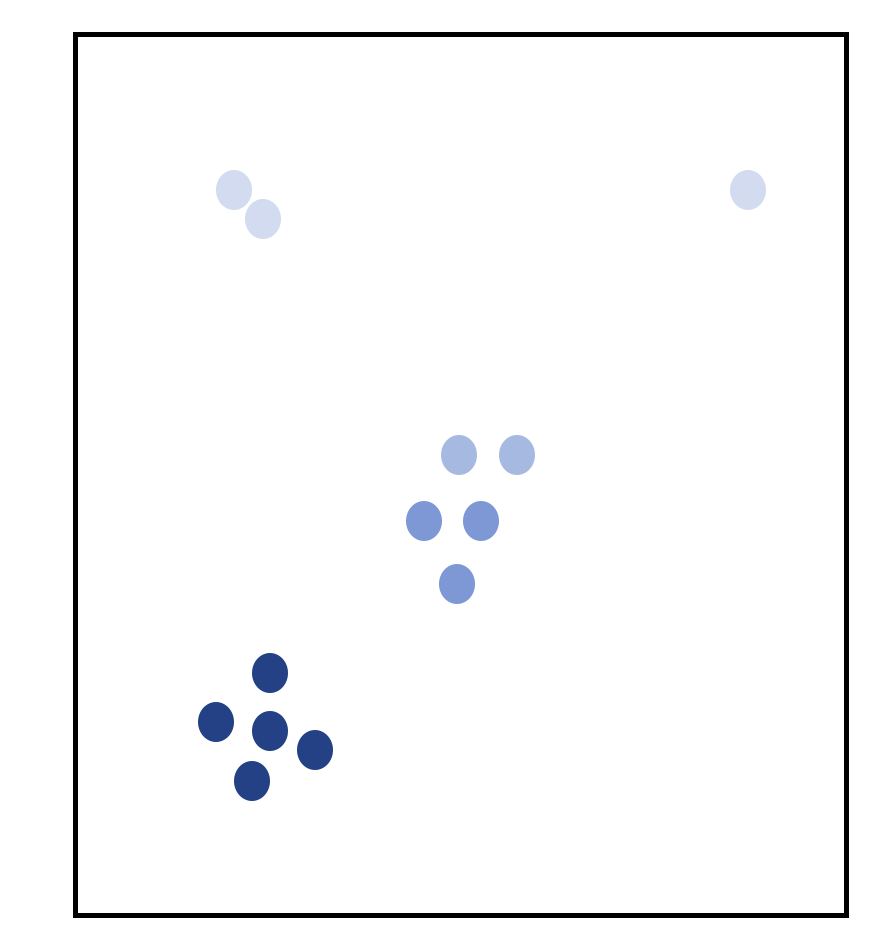}
    }\\
    \caption{A visual representation of the difference between the Contrastive and Spearman losses. The Contrastive loss considers all entities either in-cluster (blue) or out-of-cluster (orange). The Spearman loss captures the relative similarity of objects, shown in shades of blue, regardless of cluster status.}
\end{figure}

\subsubsection{Contrastive Loss}

The Siamese contrastive loss is defined as:

\noindent
\begin{equation}
\label{eq:contrastive}
L_\mathit{contrastive} = Y_\mathit{true} D + (1-Y_\mathit{true})\max(\mathit{margin}-D,0)
\end{equation}

\noindent
where $D$ is the distance between a pair of points, $Y_{true}$ is 1 if the pair contains similar objects or 0 if the objects are dissimilar, and margin is the desired separation between dissimilar objects and is a tunable hyperparameter. Equation \ref{eq:contrastive} requires that samples either belong to the same group or not, meaning that we cannot include more fine-grained similarity (e.g., these two samples are 75\% similar), in the loss function. Consequently, in this case we convert the CAPA detection sets into hard clusters with a locality sensitive hash. Employing a MinHash with one band and 64 permutations, we compute a single hash (cluster) for each binary file, where two files are similar if they lie in the same cluster ($Y_{true}=1$) and different otherwise ($Y_{true}=0$). During our experiments we employ a contrastive loss using Euclidean distance with a margin of 10, which was selected with light hyperparameter optimization on the training dataset.

\subsubsection{Spearman Loss}

While our contrastive approach assesses similarity based on CAPA clusters, it coarsely embeds binaries as ``similar" or ``not similar", when in reality some sets of CAPA labels are more similar than others. To account for finer-grained similarities, we employ a novel approach based on the Spearman rank correlation coefficient.

Specifically, advances in approximate differentiable sorting and ranking  \cite{blondel2020fast} allow us to optimize Spearman’s rank correlation coefficient with stochastic gradient descent. This allows us to compute the loss between a ground truth ranking and a predicted ranking from our model. This is desirable as it allows inserting \textit{more nuanced information into the loss function based on finer grained degree}, rather than a simple binary similar/dissimilar decision. In our experiments the ranking is based on similarity, from most similar to least. Given integer ranks we can define Spearman’s rank correlation coefficient as 

\noindent
\begin{equation}
    r = 1 - \frac{6\sum_i (R(X_i)-R(Y_i))^2}{n(n^2-1)}
\end{equation}

\noindent
where $X_i$ and $Y_i$  are the ground truth similarity and predicted similarity of data point $i$ and $R(X_i)$ and $R(Y_i)$ are the corresponding ranks. We assess ground truth similarity between two CAPA capability sets as their Jaccard similarity, and use the soft rank implementation from Blondel et al. \cite{blondel2020fast} to compare predicted and ground truth ranks. For a given batch, we use these ranks to establish the Spearman rank correlation coefficient – the loss for that batch.

\subsection{Training Process}
\label{sec:train}

All the layers of our networks are initialized by the Xavier algorithm \cite{glorot2010understanding} and trained with stochastic gradient descent (SGD) to a maximum of 30 epochs with a learning rate of 0.001. The batching algorithm used for SGD training was modified to better support our metric learning loss functions.  Ordinarily, each batch contains $C$ randomly sampled (with replacement) clusters and $M$ randomly sampled PE files from each cluster (again, sampled with replacement). For our binary similarity problem, we are confronted with two complications. First, each cluster can potentially contain both goodware and malware unlike typical metric learning problems where clusters are homogeneous. Second, we have an extremely large number of clusters ($\mathcal{O}(10^5 )$). To address the goodware and malware heterogeneity concern, we split each cluster $C$ into two clusters, $C$-goodware and $C$-malware. When we sample $C$ clusters, we do so from the combination of all $C$-goodware and $C$-malware clusters. To address the second concern, we sample $C$ clusters without replacement and define the end of the epoch when the model has processed examples from every cluster. This algorithm ensures we cover the full space of goodware, malware, and clusters in each epoch while maintaining a balance between positive and negative pairs in each batch. For these experiments, we set $C=20$ and $M=4$.

\subsection{Transfer Process}

After training the embedding, we measure the embedding's usability on various malware classification tasks. For our experiments, we train an embedding network using the EMBER 2018 training partition and extracted CAPA labels. Once we have a trained embedding network, we can use this to extract embeddings from any dataset with EMBER features. Using extracted embeddings for a given dataset, we can then fit a lightweight classifier over the embeddings and corresponding labels to make predictions for arbitrary different tasks.

The choice of the best final classifier for each task is not obvious.  Typically, generalization-based learning using ensemble methods (e.g., random forests or gradient boosted trees) provide state of the art performance on malware tasks. However, our feature space is unique in that distances between two training points have meaning and decision tree methods that rely on splitting individual features may have difficulty capturing that geometry. Notably, SVMs are a generalization-based method that could take our metric space into account, but we ignore it here due to the computational cost of training an SVM on very large datasets. An alternative would be an instance-based learning algorithm (e.g., $k$-nearest neighbors), which explicitly considers distances between training data points. As we will show in the following evaluation, we consider both instance and generalization-based classifiers, and the best classifier can vary based on the transfer task.

\section{Experiments}
\label{sec:experiments}

\subsection{Embedding Networks}

We trained various embedding networks using EMBER feature vectors and CAPA labeling extracted from PEs in the EMBER 2018 \cite{anderson2018ember} train partition. These consist of: (1) contrastive loss on CAPA clusters, (2) Spearman loss on Jaccard similarities between CAPA attribute sets, and mixed objective Spearman and Contrastive loss, where the net loss term is the sum of the losses. We also test a weighting of 10x on the Spearman loss term to bring the contributions from each constituent loss term to roughly the same order of magnitude.

% \end{itemize}

We trained each embedding network according to the procedure discussed in Section \ref{sec:train}. Since deep learning models are not amenable to convex optimization (i.e., no global minimum guarantee), we trained five different instantiations of each model in order to assess variance in performance. When performing transfer task experiments, we then aggregated mean and standard deviation statistics across embeddings from all five networks of a given type. 

\subsection{Transfer Experiments on EMBER}
\label{sec:transfer_ember}

As an initial evaluation of our embeddings, we performed two transfer tasks on the EMBER dataset: malware detection and malware family classification.

\begin{figure}[!h]
    \centering
    \subfloat[\mbox{Goodware/Malware}]{\label{fig:goodMalEMBER}
            \includegraphics[width=0.48\linewidth]{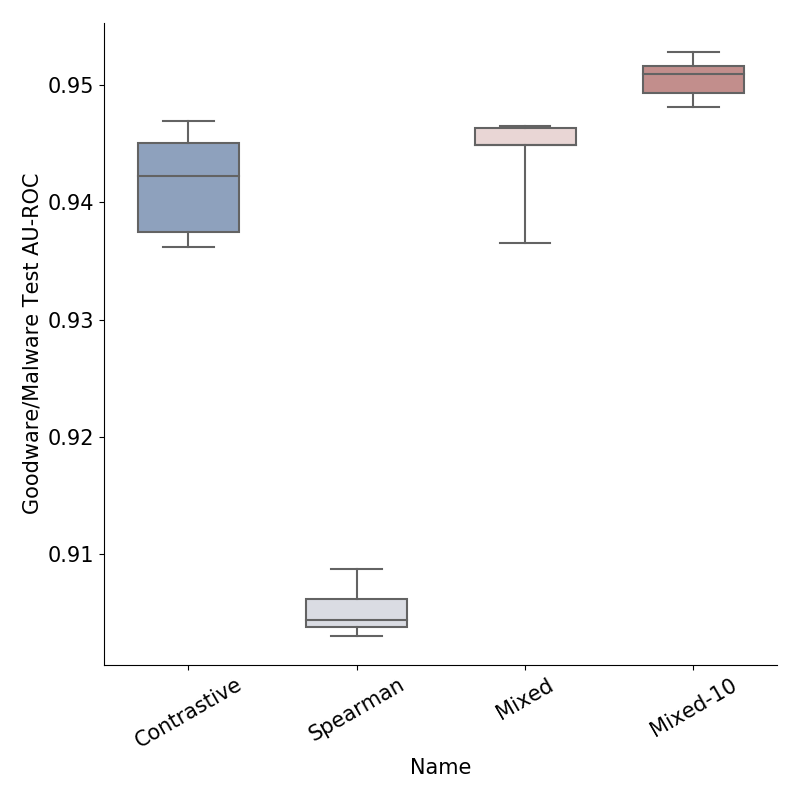}
    }
    \subfloat[Malware Family]{\label{fig:malFamilyEMBER}
            \includegraphics[width=0.48\linewidth]{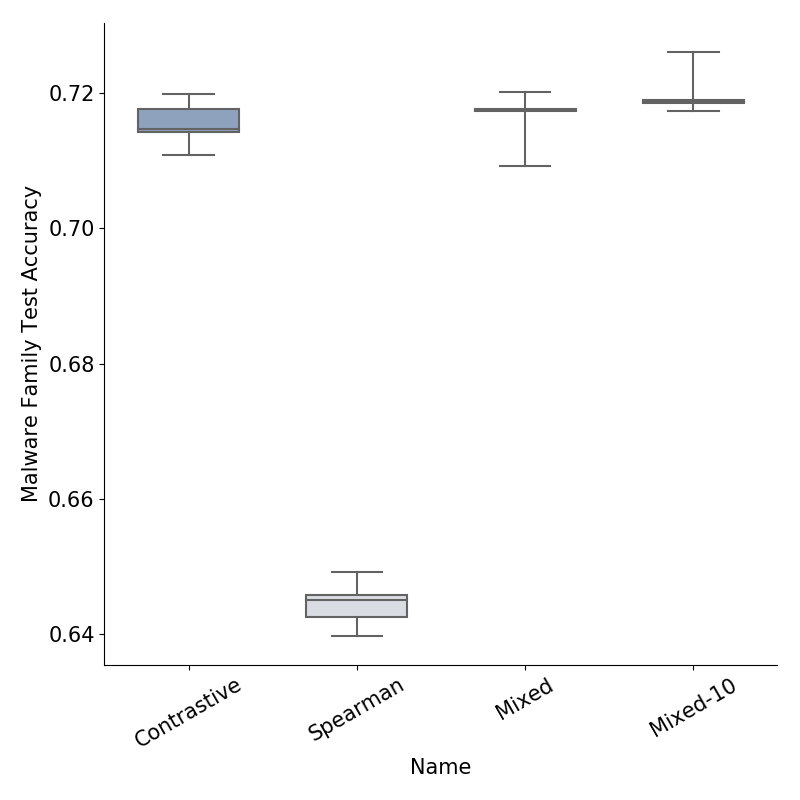}
    }\\
    \caption{Transfer Experiments on EMBER. In \protect\subref{fig:goodMalEMBER}, results of the Goodware/Malware transfer experiment are reported in terms of the area under the ROC curve (AU-ROC). In \protect\subref{fig:malFamilyEMBER}, results of the malware family transfer experiment are reported in terms of accuracy. For both experiments, classifiers trained on Spearman embeddings underperformed those trained on Contrastive embeddings while classifiers trained on embeddings derived from our weighted mixed-objective loss were the top performers.
    }
\end{figure}

The malware detection transfer task aims to detect malware using EMBER's malicious and benign labels. For this task, we extracted embeddings across both train and test partitions of EMBER 2018. We then fit a lightGBM ensemble with 1000 trees and otherwise default parameters over the embeddings extracted from the training set, and evaluated using embeddings extracted from the test set. The results of this experiment are shown in Figure \ref{fig:goodMalEMBER} in terms of the area under the ROC curve (AU-ROC) on the test set. 

In this experimental setting, we tried different weightings of the mixed objective loss, with the Spearman component both unweighted and up-weighted by a factor of 10 to be on the same scale as the contrastive component. We notice that the transfer performance on the contrastive loss embedding significantly outperforms the transfer performance on the Spearman loss embedding. However, both embeddings which use a combination of the two losses offer better classification performance than strictly either of the embeddings trained on a solo loss (Spearman or Contrastive), with the weighted mixed objective loss outperforming the unweighted. Note that none of the transfer malicious/benign classifiers on EMBER 2018 exceed the baseline model from \cite{anderson2018ember}.

The second transfer task is a malware family recognition task, which utilizes a 1-nearest neighbor classifier in the embedding space in conjunction with the EMBER 2018 malware family labels (derived via AVClass \cite{sebastian2016avclass}) to predict the family for malicious samples. These results are shown in Figure \ref{fig:malFamilyEMBER}. While here the performance evaluation is in terms of accuracy not AU-ROC, we notice the same performance trend across embedding types as for the detection transfer task -- that is, the mixed objective equals or out performs contrastive or Spearman loss in isolation.

\subsection{Transfer Experiments on SOREL-20M}

We additionally evaluated the performance of our embeddings for different tasks on the SOREL-20M dataset. SOREL is a large industry-scale dataset with publicly available labeling telemetry beyond just malicious/benign detection; it also contains public labeling telemetry for 11 distinct malware attributes, namely: Adware, Crypto Miner, Downloader, Dropper, File Infector, Flooder, Installer, Packed, Ransomware, Spyware, and Worm. Note that these attributes are non-mutually exclusive across samples, meaning that a given malware sample can have multiple malware attributes. We can think of the SOREL's malware attributes as defining high-level behaviors encapsulate fine-grained capabilities, similar to those identified by CAPA.

\begin{figure}[!h]
    \centering
    \includegraphics[width=0.66\linewidth]{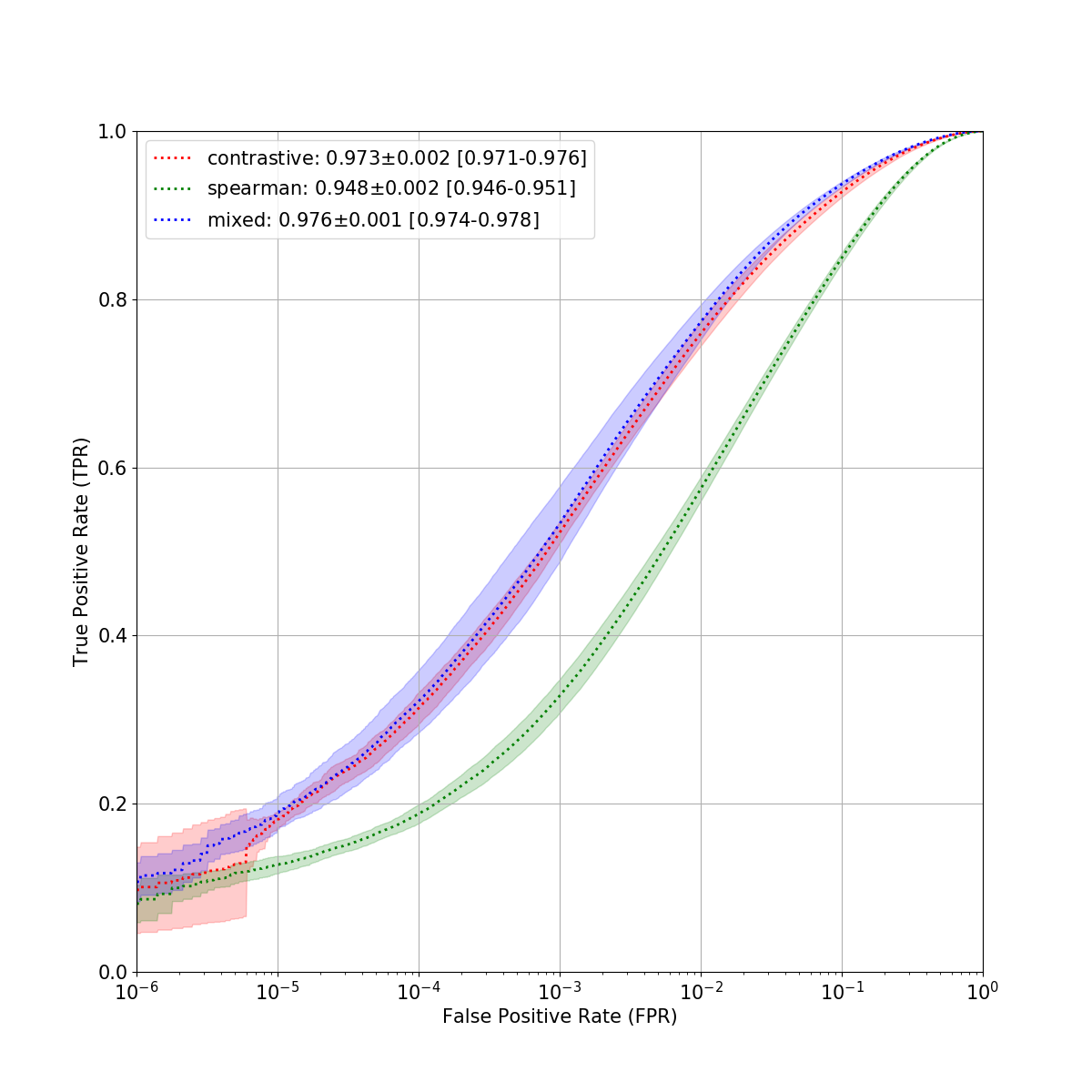}
    %\caption{Detection transfer experiment on SOREL-20M.}
    \caption{Results of the Detection Transfer Experiment on SOREL-20M. Classifiers trained on the Contrastive and Mixed embeddings have the highest AU-ROCs of $0.973 \pm 0.002$ and $0.976 \pm 0.001$, respectively. Classifiers trained on the Spearman embedding have an AU-ROC of $0.948 \pm  0.002$.}
    \label{fig:goodMalEMBER_SOREL_ROCs}
\end{figure}

SOREL also contains different data with a different data distribution than EMBER (on which the embeddings were extracted). This suggests any strong performance over the EMBER-to-SOREL transition is a  indication of the robustness of our approach to producing general purpose representations for downstream tasks. We extracted 32-dimensional embeddings for all of the SOREL samples apriori, and trained task-specific lightGBM classifiers on the extracted embeddings. We assessed embedding performance/quality for two distinct tasks on SOREL: malware detection and malware attribute labeling. 

Results from the malware detection task are shown in Figure \ref{fig:goodMalEMBER_SOREL_ROCs}. Embedding extraction and lightGBM training was performed fives times to obtain error bars. Consistent with our transfer experiments on EMBER, the mixed objective embedding yields the highest AU-ROC, slightly outperforming the contrastive embedding and significantly outperforming the Spearman embedding. For reference, the lightGBM baseline trained on the full 2381-dimensional features has an AU-ROC of $0.981 \pm 0.002$ \cite{harang2020sorel}, a relative average improvement over the top-performing mixed objective model of 0.05\%. However, storing the full 2381-dimensional feature vectors requires 74.4 times the amount of storage of as that of the 32-dimensional embeddings, indicating that in practice, the top-performing embeddings significantly reduce the storage burden at a slight reduction in net performance.

Our results from the malware attribute labeling task are shown in Table \ref{tab:attributes}. For this task, we fit lightGBM classifiers across each of the malware tags, using 1-hit for each tag as a criterion for presence of the attribute, consistent with Harang and Rudd \cite{harang2020sorel}. Notably, we see a similar pattern with the \textit{Mixed-10} loss on average outperforming the contrastive loss and both losses on average, outperforming the Spearman loss. Note however, that the mixed loss under-performs the contrastive loss when Spearman performs especially poorly. On average, the results for tagging under-perform the baseline provided with SOREL-20M benchmark, though this is a somewhat invalid comparison, as the attribute baselines from Harang and Rudd \cite{harang2020sorel} utilized a large multi-target network, factoring in number of vendor hits, malicious/benign classification, and simultaneous attribute predictions; thus some of the performance discrepancy is likely due to limitations of single-target classifiers. 

\begin{table}[!h]\centering
\begin{tabular}{c|c|c|c}
     & Contrastive & Spearman & Mixed  \\ \hline
    Adware & $\bf0.917 \pm 0.005$ & $0.883 \pm 0.005$ & $\bf 0.917 \pm 0.002$ \\ \hline
    Crypto Miner & $\bf 0.976 \pm 0.004$ & $0.962 \pm 0.001$ & $\bf 0.976 \pm 0.003$ \\ \hline
    Downloader & $0.832 \pm 0.007$ & $0.798 \pm 0.005$ & $\bf 0.835 \pm 0.004$ \\ \hline
    Dropper & $0.819 \pm 0.009$ & $0.773 \pm 0.005$ & $\bf 0.824 \pm 0.011$\\ \hline
    File Infector & $0.878 \pm 0.003$ & $0.834 \pm 0.005$ & $\bf 0.885 \pm 0.007$\\ \hline
    Flooder & $\bf 0.982 \pm 0.006$ & $0.981 \pm 0.003$ & $0.979 \pm 0.003$ \\ \hline
    Installer & $0.957 \pm 0.003$ & $0.929 \pm 0.002$ & $\bf 0.962 \pm 0.002$\\ \hline
    Packed & $\bf0.783 \pm 0.003$  & $0.742 \pm 0.004$ & $0.779 \pm 0.013$\\ \hline
    Ransomware & $0.977 \pm 0.003$ & $0.959 \pm 0.002$ & $\bf 0.978 \pm 0.003$\\ \hline
    Spyware & $\bf 0.848 \pm 0.010$ & $0.776 \pm 0.003$ & $0.846 \pm 0.014$\\ \hline
    Worm & $\bf 0.877 \pm 0.014$ & $0.804 \pm 0.014$ & $\bf 0.877 \pm 0.014$\\ \hline
\end{tabular}
  \caption{Results from the malware attribute transfer experiment on SOREL-20M. Mean AU-ROC and AU-ROC standard deviation are reported, with results aggregated over five runs. Best results are shown in bold.}
  \label{tab:attributes}
\end{table}

\section{Practical Deployment Considerations}

\subsection{Enriching Metric Embeddings with Common Transfer Tasks}
\label{sec:binary_enrichment}

\begin{figure*}[!h]
    \centering
    \includegraphics[width=\textwidth]{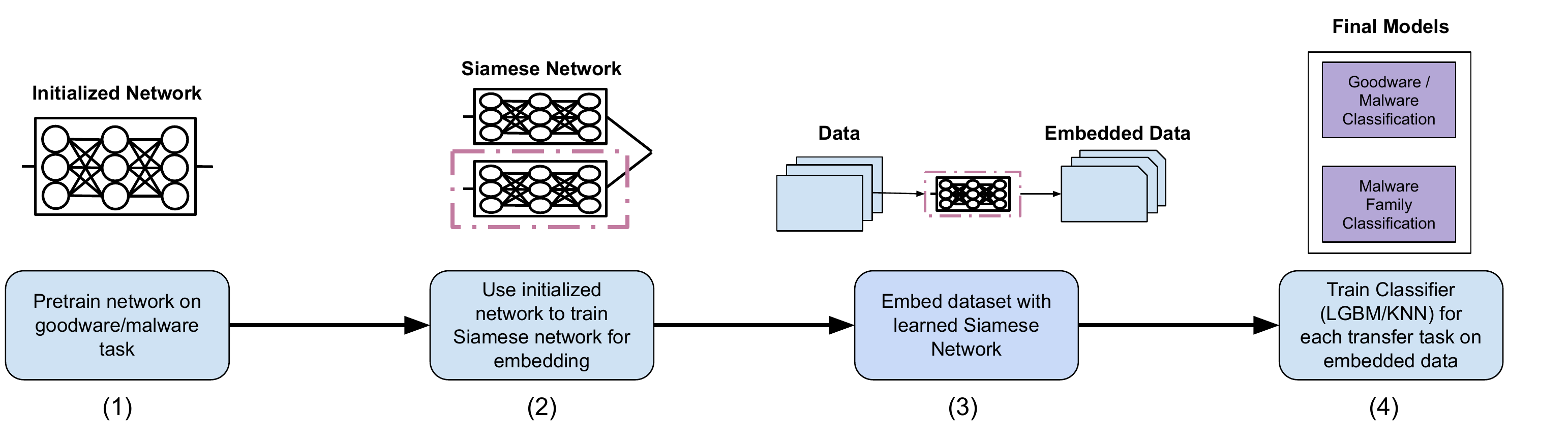}
    \caption{This figure depicts our training process for enriching our metric embeddings with a detection task. (1) We pretrain a network
        on goodware/malware classification to initialize effective weights for the next task.
        (2) We use the pretrained network as the base for an embedding network, which we train using a single-objective (contrastive) loss or multi-objective (contrastive and a binary cross-entropy detection) loss. (3) We then embed a dataset using the learned embedding from (2), represented by the dashed purple box. (4) We perform and evaluate transfer tasks (goodware/malware classification and malware family/type classification).}
    \label{fig:pipeline}
\end{figure*}

In prior sections, we maintained intentional separation between metric learning tasks and downstream transfer tasks in order to assess different metric learning approaches. In practical applications, where we prioritize downstream task performance, we can additionally incorporate these tasks when training the metric learning. Methods of combining downstream transfer tasks with metric learning include pretraining on the transfer task prior to metric learning and incorporating the transfer task as an additional objective while performing metric learning. Similar approaches have been foundational within applied computer vision and facial recognition~\cite{schroff2015facenet,donahue2014decaf,rudd2016moon}.
In this section, we ask: can enriching our metric embeddings with a detection task lead to superior performance during downstream transfer? To answer this, we repeat a variation of our transfer experiments on EMBER from Sec.~\ref{sec:transfer_ember}. These experimental regimes are depicted in Fig~\ref{fig:pipeline}.

\begin{figure}[!h]
    \centering
    \subfloat[\mbox{Goodware/Malware}]{\label{fig:goodMalEMBERTransfer}
            \includegraphics[width=0.48\linewidth]{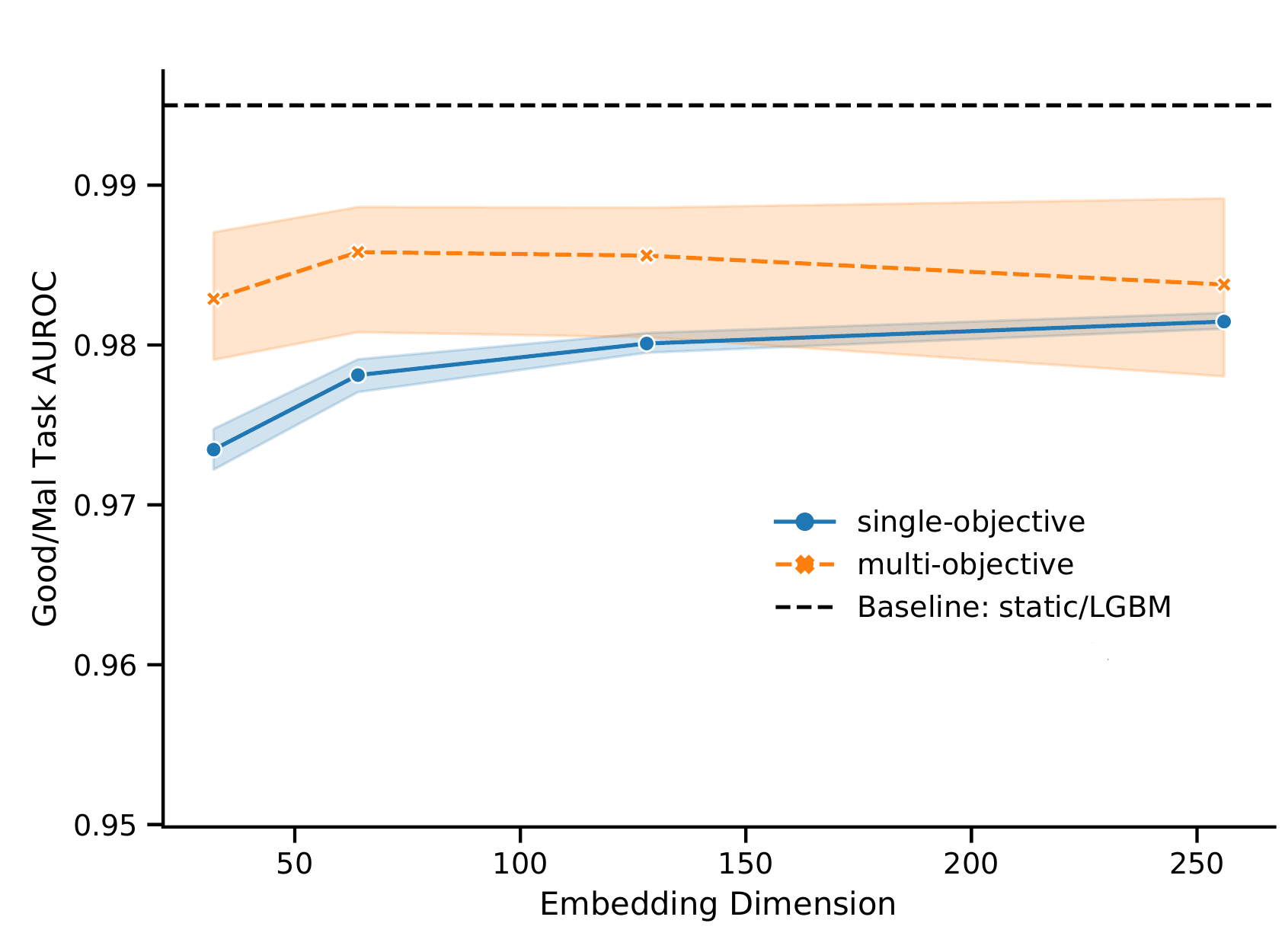}
    }
    \subfloat[Malware Family]{\label{fig:malFamilyEMBERTransfer}
            \includegraphics[width=0.48\linewidth]{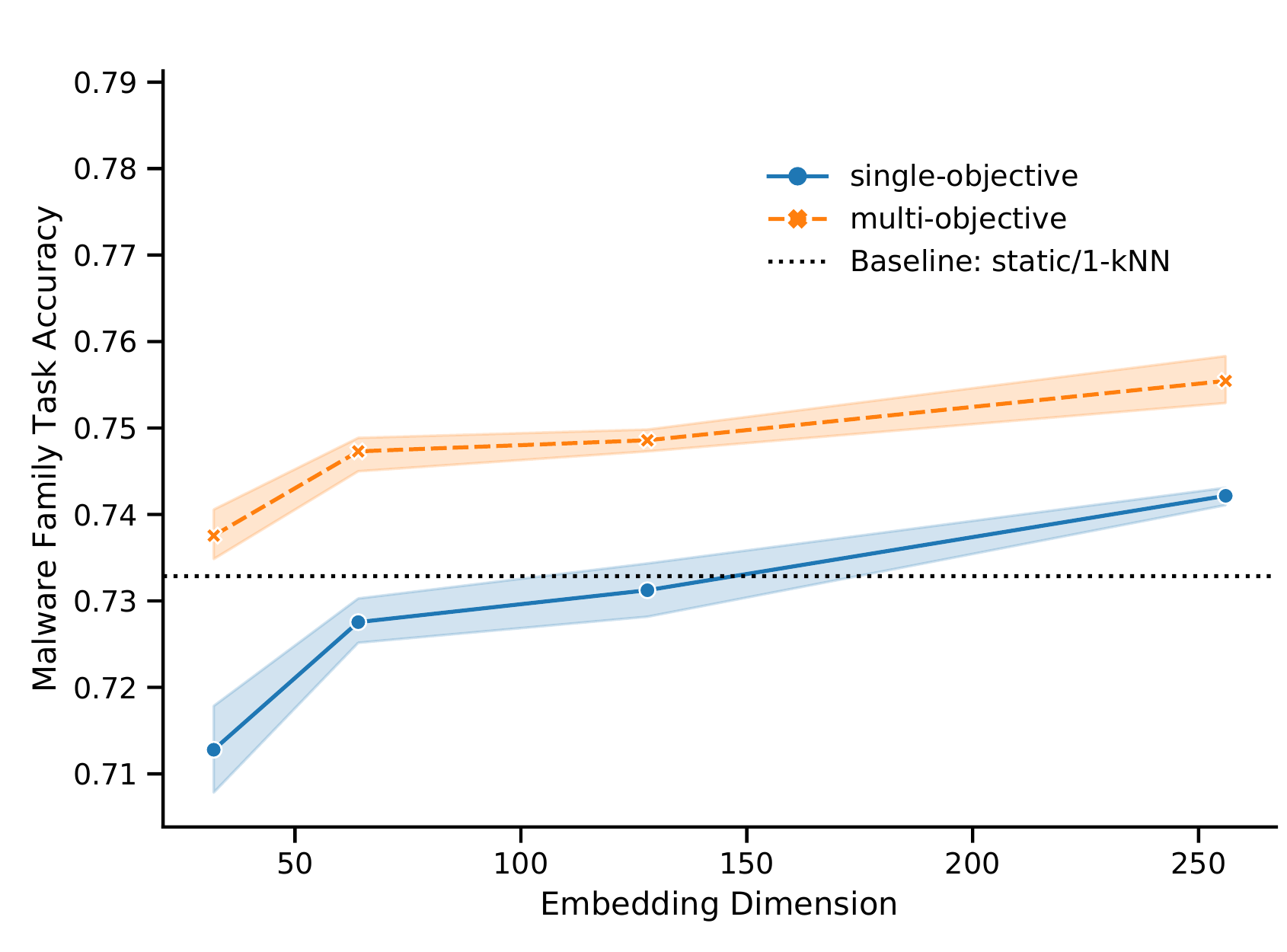}
    }\\
    \caption{Transfer Experiments on EMBER. In \protect\subref{fig:goodMalEMBERTransfer}, results of the goodware/malware transfer experiment are reported in terms of the area under the ROC curve (AU-ROC). In \protect\subref{fig:malFamilyEMBERTransfer}, results of the malware family transfer experiment are reported in terms of accuracy. For both experiments, classifiers trained on Spearman embeddings underperformed those trained on Contrastive embeddings while classifiers trained on embeddings derived from our weighted mixed-objective loss were the top performers.
    \label{fig:enrichment}
    }
\end{figure}

For our embedding network, we used the same architecture as shown in Fig.~\ref{fig:net_architecture}, though this time we vary the size of the final embedding layer to study the effects of different embedding sizes. For pretraining, we use the EMBER 2018 training set with malicious/benign labels, a final dense layer to reduce to 1D, and a sigmoid activation function. We train the initial representation using a binary cross entropy loss. The learned weights are used to initialize the embedding network. We then train the embedding network using a contrastive loss on clusters, as well as an optional malicious/benign loss on the EMBER train set labels. Note that for these experiments, we slightly modify our clustering methodology using VirusTotal's vhash instead of CAPA, with approaches to cluster sampling remaining  consistent. We then use the learned embedding network to extract embeddings over the EMBER training set. Finally, we use a LightGBM model to transfer to the specific downstream task. Results of these experiments are shown in Fig.~\ref{fig:enrichment}.

 In both experimental regimes, we tried four different embedding dimensions: 32, 64, 128, and 256. For both tasks, we found that increasing the embedding dimension from 32 to 64 noticeably enhanced performance, regardless of how the embeddings were derived. For goodware/malware classification, when the embedding is derived only from a contrastive loss over clusters, performance increased monotonically with added embedding dimension although gains were gradual beyond 64 dimensions. We also note a slight reduction in variance between runs. When the embeddings are derived from both a contrastive loss over clusters and a classification loss over labels (the multi-objective regime), increasing beyond 64 dimensions decreased performance and increased variance. For the malware family classification task, we also noticed a monotonic increase in performance over both embedding types with embedding dimension, with the steepest performance increase between 32 and 64 dimensions.  

 The goodware/malware classification experiments showed that both the contrastive (single-objective) embedding and the multi-objective embeddings offered performance superior to that of the neural network used to initialize the embedding weights, but fell short of a baseline LightGBM classifier trained directly on EMBER features. Meanwhile, in the malware family classification task, we find that regardless of embedding dimension, the multi-objective embedding offers superior performance to the baseline model trained on the much higher-dimensional EMBER feature space, and even the single-objective contrastive embedding outperforms the EMBER feature space with sufficient increase in dimensionality.  

\subsection{Qualitative Visualization}

\begin{figure}[!h]
    \centering
    \includegraphics[width=\linewidth]{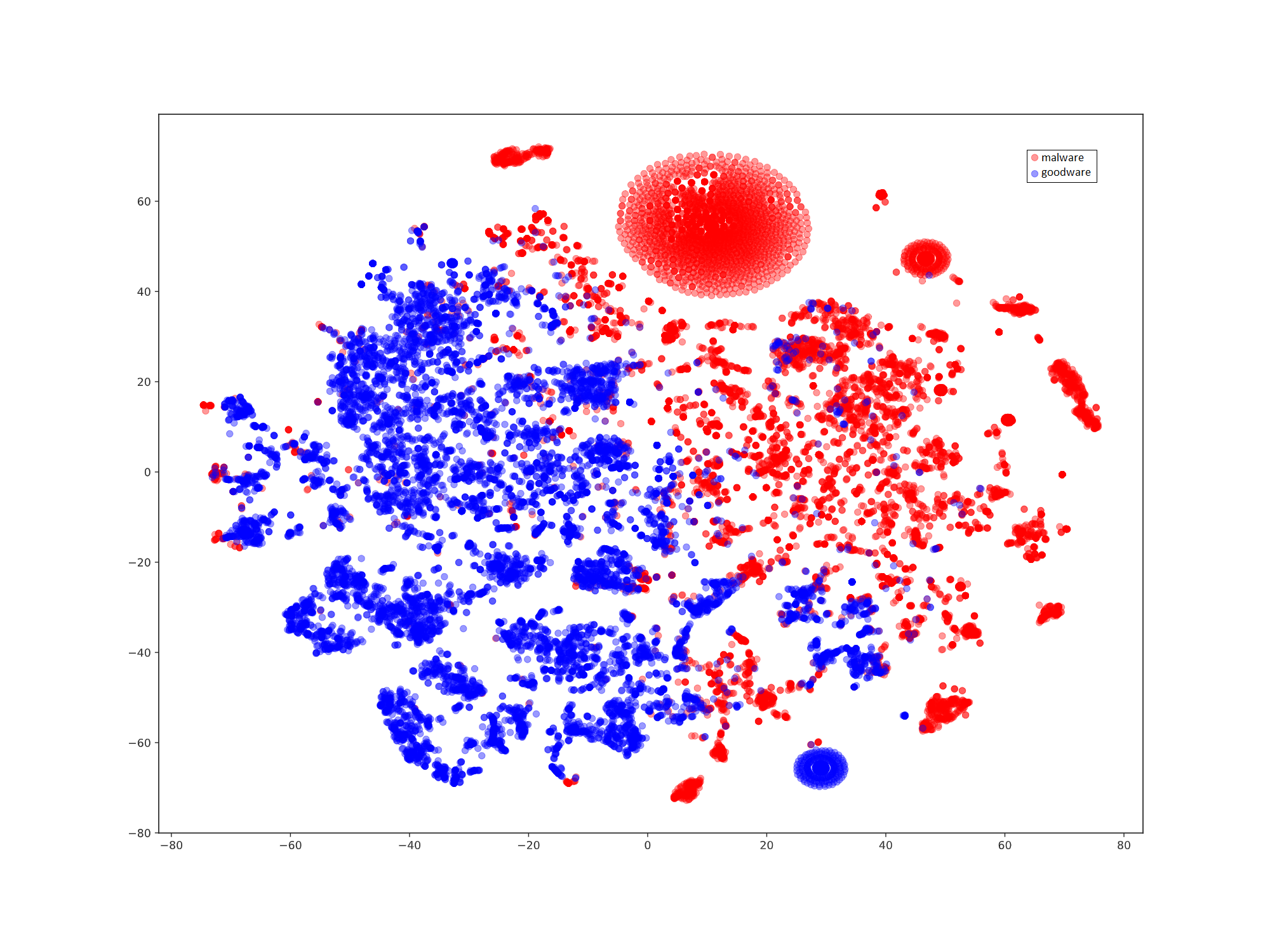}
    \caption{A qualitative t-SNE visualization of embedded samples from the EMBER test set corresponding to goodware/malware labels. The 64D multi-objective embedding was used to generate these samples. An equivalent number of samples were randomly selected from goodware/malware classes.}
    \label{fig:goodMalTSNE}
\end{figure}

Our quantitative results from Sec.~\ref{sec:binary_enrichment} suggest that our learned embeddings provide separability between classes for a variety of tasks, and in some cases even provide improved  discriminability over the much higher dimension (i.e., 2,381 features) EMBER feature space. In this section, we examine qualitative aspects of these learned embeddings through t-distributed stochastic neighbor embedding (t-SNE) visualizations~\cite{van2008visualizing}. We generate 2D t-SNE visualizations of sample embeddings from the EMBER 2018 test set, sampling over different labels/types. Our base sample embeddings were generated from the multi-dimensional 64-dimensional network described in Sec.~\ref{sec:binary_enrichment}, since they showed superior performance in the transfer tasks evaluated.

A t-SNE visualization of randomly sampled goodware/malware embeddings is shown in Fig. \ref{fig:goodMalTSNE}, with equal sampling over goodware and malware classes. Note that at a high level, we see general separability between the goodware and malware samples. Based on performance, many of the samples of heterogeneous class that do not appear separated in the 2D projected space are still likely separated in the full 64D embedding space, though these separations would naturally be much more subtle and nonlinear. Specifically of note, in the 2D projected space, we see clear evidence of a decision boundary yet simultaneously a formation of dense clusters of samples, indicating that both components of the multi-objective loss appear to have an influence on the embedded space. The contrastive component with respect to clusters creates distinct groups of sample points, while the binary cross-entropy component separates malware from goodware. 

\begin{figure}[!h]
    \centering
    \includegraphics[width=\linewidth]{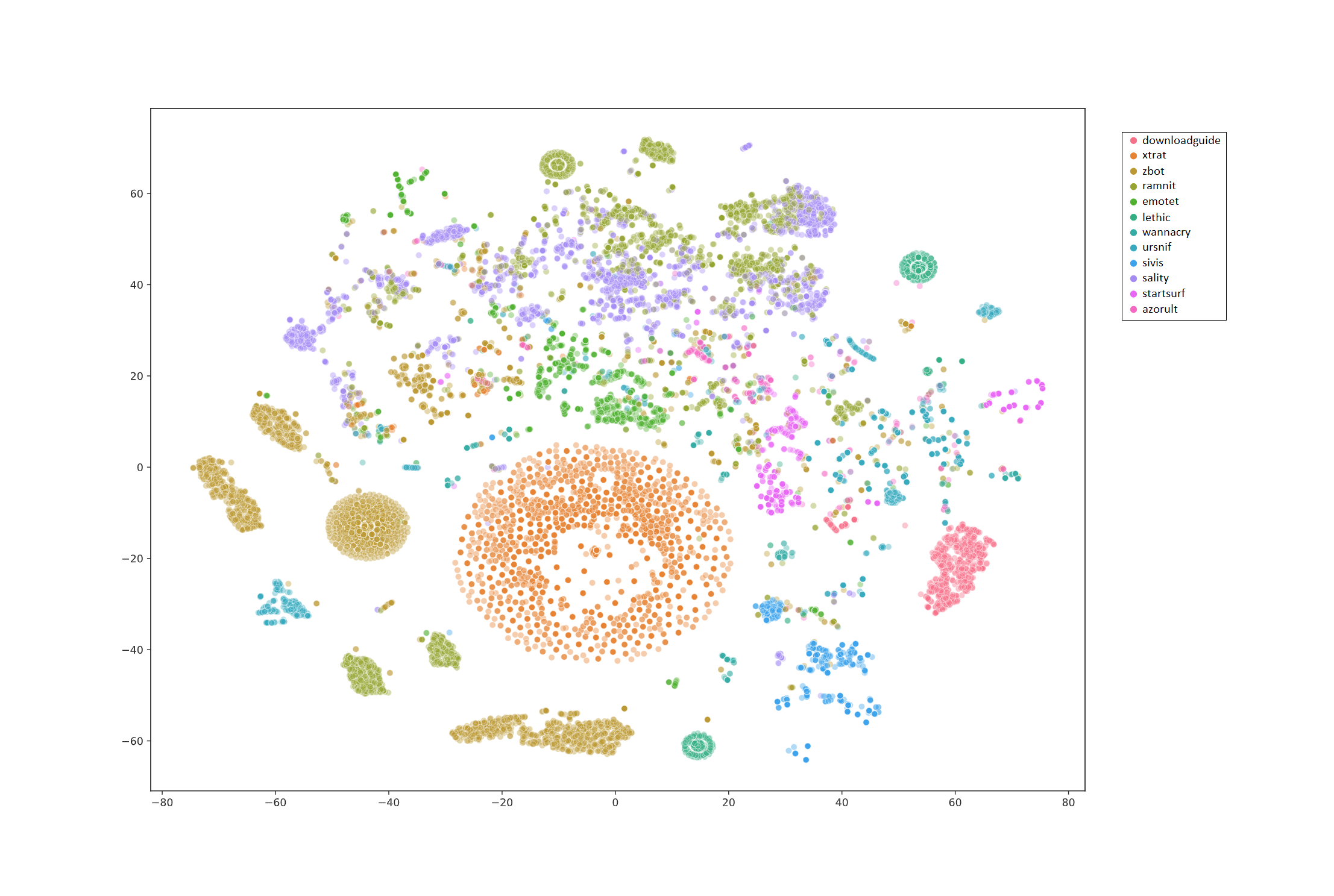}
    \caption{A qualitative t-SNE visualization of embedded samples from the EMBER test set corresponding to 12 common malware family types. }
    \label{fig:malFamilyTSNE}
\end{figure}

In Fig.~\ref{fig:malFamilyTSNE}, we select and visualize samples corresponding to 12 heterogeneous and common malware families from the EMBER test set. Intriguingly, we see that many of these families are distinctly clustered in the t-SNE projected space, even though family information was not used to learn the embedding space. This demonstrates the efficacy of the contrastive loss in learning behavioral clusters which can be applied towards alternative downstream transfer tasks beyond those directly included in the objective function.

\subsection{Analysis of Adversarial Robustness}
Given the adversarial nature of our transfer tasks, it is natural for us to investigate the robustness of our learned embeddings to evasion attacks applied in those settings. While many adversarial evasion attacks have been formulated against malware classifiers \cite{kolosnjaji2018adversarial,suciu2019exploring,anderson2018learning,song2022mabmalware}, we focus our analysis on realistic black box attacks leveraging evolutionary learning techniques \cite{demetrio2021adversarial,demetrio2021functionality}. These attacks are applicable to both traditional ML models and the neural network models proposed in this paper, making them ideal for comparing robustness in a consistent way. Furthermore, since they manipulate the binaries directly, the attacks produce {\it feasible} adversarial examples that ensure the resultant malware binary is still fully functional. By comparison, white box attacks or those that operate in the feature space require the adversary to gain extraordinary access to the classifier and its gradients, or additional steps to translate feature space manipulations back into the original problem space to create the functional binary \cite{pierazzi2020intriguing}.

Specifically, we use Demetrio et al.'s GAMMA evolutionary framework \cite{demetrio2021adversarial, demetrio2021functionality} as implemented in the SecML-Malware framework\footnote{\url{https://github.com/pralab/secml_malware}} to examine robustness of our embedding models and compare it against a baseline LightGBM model trained directly on the 2,381 EMBER features. Two specific attack formulations are presented: Section Injection and Content Shifting. In the Section Injection attack, the adversary adds new section content taken from known-good binaries into the existing malware binary in a way that does not change the other sections. Meanwhile, the Content Shifting attack adds padding from goodware to shift the offset of latter sections within the binary and change their location relative to other structures, such as headers. The evolutionary framework chooses the manipulations evoking the largest reduction in classifier confidence at each attack iteration and uses them as the basis for further manipulation in the next round. In our experiments, we allow up to 50 attack iterations and measure adversary success as the reduction in detection rate from the respective model baselines.

\begin{figure}[!h]
    \centering
    \subfloat[Section Injection Attack]{\label{fig:section_injection}
            \includegraphics[width=0.48\linewidth]{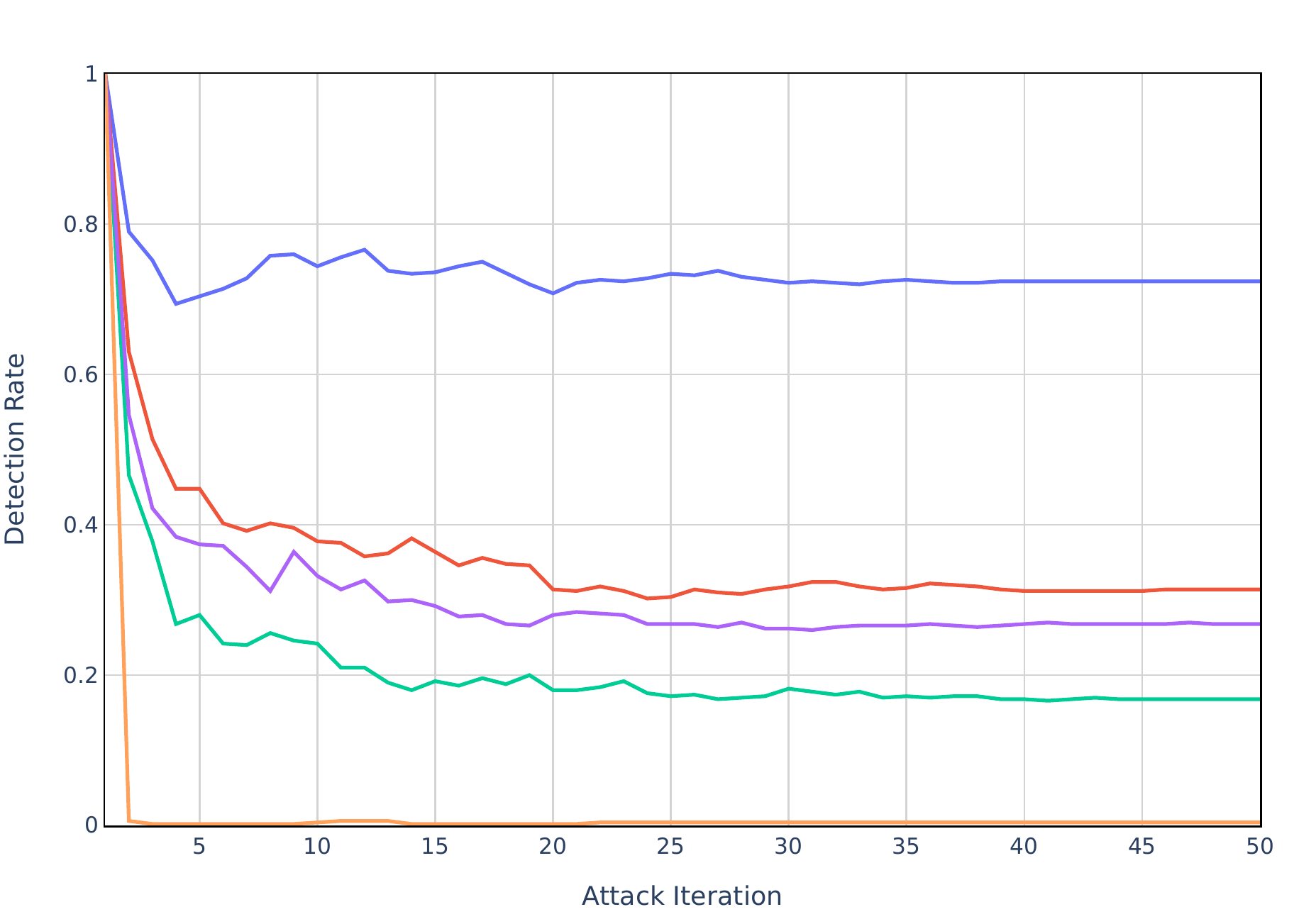}
    }
    \subfloat[Content Shifting Attack]
    {\label{fig:content_shift}
            \includegraphics[width=0.48\linewidth]{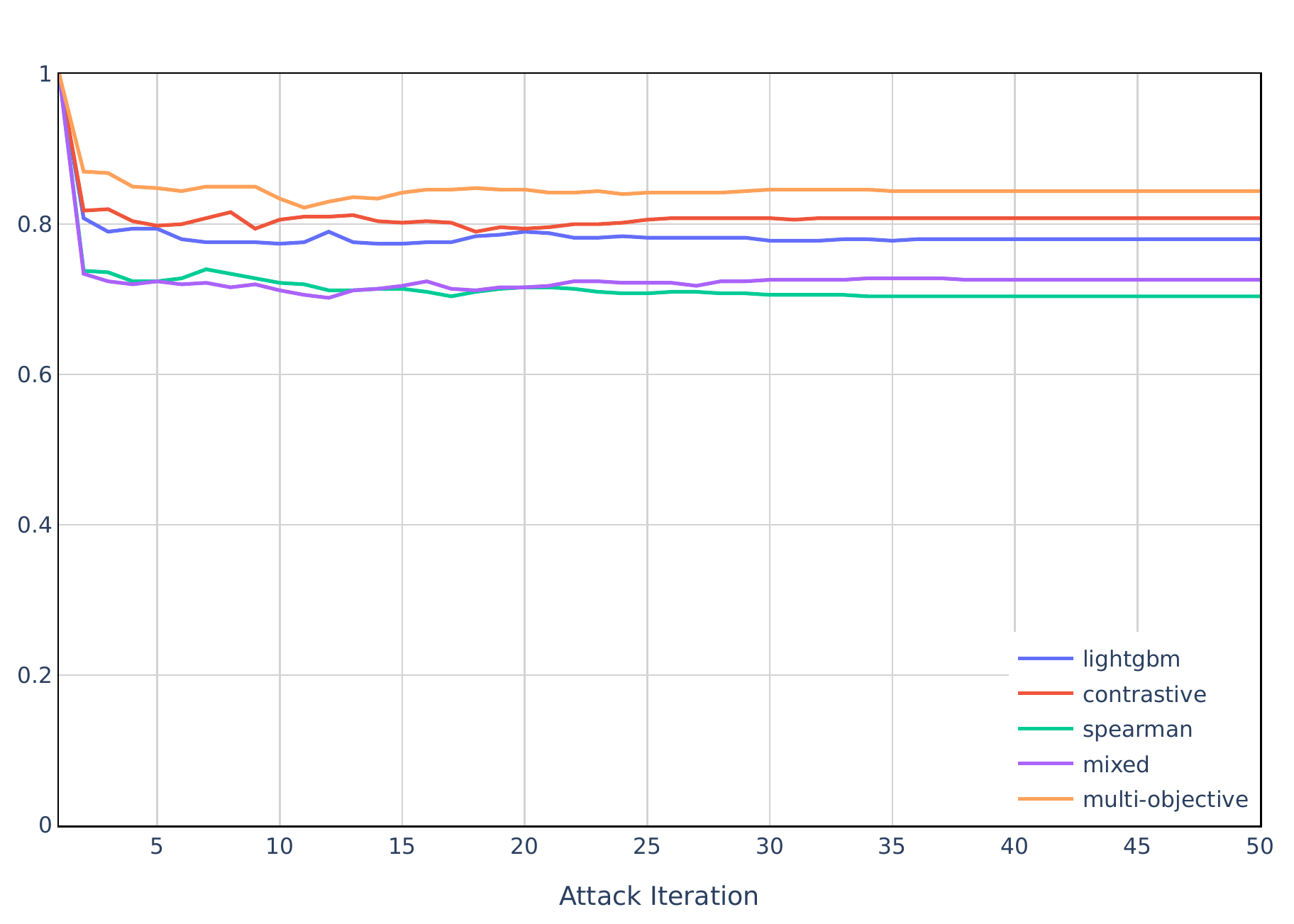}
    }\\
    \caption{Detection rates of embedding networks and LightGBM baseline under black-box evasion attack using the GAMMA genetic optimization framework \cite{demetrio2021adversarial,demetrio2021functionality}.}
    \label{fig:adversarial_experiments}
\end{figure}

The results of our experiments, shown in Figure \ref{fig:adversarial_experiments}, demonstrate some interesting trends when examining the success of the GAMMA attack against our embedding models compared to the LightGBM classifier trained directly on the static features. First, it is clear that our embedding models are more sensitive to the Section Injection attack than the LightGBM baseline. The attack reduces the detection rate for our models to between 16.8 and 30.8\% for our standard embedding models, and causes complete evasion of the multi-objective model trained on the malware classification task (i.e., detection rate of 0\%). At the same time, some embedding models outperform the baseline versus the Content Shifting attack, maintaining a detection rate of up to 84.6\% versus 78.2\% for the LightGBM model. In both cases, the contrastive-only embedding model performs well compared to the other embedding model variants, while Spearman generally performs poorly.

While a complete analysis of the underlying reasons for these results are beyond the scope of this paper, we can hypothesize that the Content Shifting attack has limited impact across all evaluated models because the manipulation does not significantly change the static features underlying the models, or only changes those features that are relatively unimportant in the final model output. The Section Injection attack, on the other hand, affects features that the embedding models rely heavily on for separating samples within the embedding space, such as entropy and number of sections. One interesting observation is that the contrastive-only model remains robust relative to the other embedding models, and that could be due to the coarse nature of the objective itself -- focusing on broad notions of positive and negative examples. When losses with additional granularity are used, such as Spearman loss, robustness is noticeably reduced. The binary cross-entropy objective of the multi-objective model, meanwhile, interacts with the metric embedding in a way that makes the decision boundary extremely brittle w.r.t. specific features, as observed in Figure \ref{fig:goodMalTSNE}. Overall, however, it seems like objectives that enforce additional separation among examples within the broader goodware or malware classes open the possibility of introducing weaknesses, perhaps by becoming overly reliant on a small number of features to achieve that separation.
\section{Discussion}

In this paper, we have introduced two different approaches to enrich metric embeddings with static disassembly capabilities information and performed evaluations thereof on multiple downstream tasks. These approaches, outlined in Sec.~\ref{sec:approach} consist of a fine-grained Spearman embedding approach and a coarse-grained contrastive embedding approach. In the vast majority of our experiments in Sec.~\ref{sec:experiments}, the coarse-grained contrastive approach exhibited superior performance to the finer-grained Spearman approach. In some respects, this is not surprising, as contrastive loss inherently forces separability in a way that the Spearman loss does not. An in-depth examination of similarity distributions and adoption of additional similarity measures other than Jaccard similarity could be helpful in improving the Spearman embedding. Consistent with other literature in the ML security/applied ML space, we found that combining both Spearman and contrastive embedding losses generally improved performance as did balancing loss contributions to a similar order of magnitude \cite{rudd2019aloha,rudd2022transformers,rudd2016moon}. Furthermore, adding task-specific objectives greatly improved performance on multiple downstream tasks, but at the cost of catastrophic loss of robustness to evasion.

When trained from scratch, using no transfer learning or auxiliary downstream task losses, our embeddings performed comparably to classifiers trained on raw features for certain tasks, but did not did not work so well for others. Among a variety of factors, this may be due to including semantic information inherent to the CAPA embeddings, over/under-fitting, and hyperparameter selection. While we were not able to outperform our baselines using transfer from ``from-scratch" metric learnt embeddings alone, we were able to do so on the malware family classification task in Sec.~\ref{sec:binary_enrichment}, using heterogeneous labeling in conjunction with metric learning in a manner very similar to approaches previously explored in computer vision literature~\cite{schroff2015facenet,donahue2014decaf}. Generally, we surmise that improving performance of metric embeddings for additional tasks is trivially feasible by utilizing additional training objectives. These may include malicious/benign labels, malware attribute tags, MITRE ATT\&CK tactics (notably, CAPA outputs these as well as attributes) and additional metadata. Moreover, examining how embedding performance scales when training on larger more heterogeneous groups of samples and evaluating on substantially concept-drifted data could offer further insight into embedding performance and design (e.g., \cite{bodmas}).

Notably, a low-dimensional embedding which performs well for a variety of classification and/or information retrieval tasks could yield significant computational and storage savings over utilizing raw features or binaries. Such embeddings could be utilized in both academic contexts, where compute resources are often limited or in commercial contexts for rapid prototyping. As a reference, the SOREL-20M dataset is an order of magnitude smaller than industry datasets which are typically used to train commercial PE malware detectors, yet it still comes with a warning about potentially incurring bandwidth fees or exhausting disk space. Even in featurized format as 32-bit floating point, the SOREL-20M dataset requires 172 GB of storage. Using the embeddings introduced in this paper, this can be compressed to roughly 2.3 GB, which is small enough to fit in memory, even for most laptops.

\bibliographystyle{abbrvnat}    
\bibliography{citations}

\end{document}